%% file: acl_latex.tex
\pdfoutput=1

\documentclass[11pt]{article}

\usepackage{enumitem}
\usepackage{graphicx}
\usepackage{float}
\usepackage[export]{adjustbox} 

\usepackage{amsmath,amsthm}

\usepackage{multirow}
\usepackage[table,xcdraw]{xcolor}
\usepackage{booktabs}

\usepackage{makecell}

\usepackage{siunitx}
\usepackage{mhchem}

\usepackage{adjustbox}

\usepackage[final]{acl}

\usepackage{times}
\usepackage{latexsym}

\usepackage[T1]{fontenc}

\usepackage[utf8]{inputenc}

\usepackage{microtype}

\usepackage{inconsolata}

\usepackage[capitalise]{cleveref}

\newcommand*\samethanks[1][\value{footnote}]{\footnotemark[#1]}
\title{How Far Are LLMs from Believable AI? 
A Benchmark for Evaluating the Believability of Human Behavior Simulation}

\author{Yang Xiao$^{1}$ \quad Yi Cheng$^{1}$ \quad Jinlan Fu$^{2}$ \\
\textbf{Jiashuo Wang}$^{1}$ \quad \textbf{Wenjie Li}$^{1}$\thanks{\,\, Corresponding Authors} \quad \textbf{Pengfei Liu}$^{3}$\samethanks \\
$^{1}$The Hong Kong Polytechnic University \quad $^{2}$National University of Singapore \\ \quad $^{3}$Shanghai Jiao Tong University \\
{\small{\texttt{\{yang-alan.xiao,alyssa.cheng\}@connect.polyu.hk} \quad \texttt{jinlanjonna@gmail.com}}}\\
{\small {\texttt{csjwang@comp.polyu.edu.hk} \quad \texttt{cswjli@comp.polyu.edu.hk} \quad \texttt{pengfei@sjtu.edu.cn}}}\\
  }

\begin{document}
\maketitle
\begin{abstract}
\input{section/0Abstract}
\end{abstract}
\input{section/1Introduction}
\input{section/2Relatedworks}
\input{section/3BenchmarkConstruction}

\input{section/4Methodology}
\input{section/5Evaluate}

\input{section/6Analyses}
\input{section/7Conclusion}

\bibliography{custom}
\clearpage
\appendix
\input{section/Appendix}

\end{document}

%% file: section/0Abstract.tex
In recent years, AI has demonstrated remarkable capabilities in simulating human behaviors, particularly those implemented with large language models (LLMs). 
However, due to the lack of systematic evaluation of LLMs' simulated behaviors, the \emph{believability} of LLMs among humans remains ambiguous, i.e., it is unclear which behaviors of LLMs are convincingly human-like and which need further improvements. In this work, we design \emph{SimulateBench} to evaluate the believability of LLMs when simulating human behaviors. In specific, we evaluate the believability of LLMs based on two critical dimensions: 1) \emph{consistency}: the extent to which LLMs can behave consistently with the given information of a human to simulate; and
2) \emph{robustness}: the ability of LLMs' simulated behaviors to remain robust when faced with perturbations. SimulateBench includes 65 character profiles and a total of 8,400 questions to examine LLMs' simulated behaviors. Based on SimulateBench, we evaluate the performances of 10 widely used LLMs when simulating characters. The experimental results reveal that current LLMs struggle to align their behaviors with assigned characters and are vulnerable to perturbations in certain factors.
\footnote{Code and SimulateBench are available at an \href{https://github.com/LLMConference/emnlp_conference_2024}{anonymous GitHub repository.}}

%% file: section/1Introduction.tex
\section{Introduction}
AI has shown promise to simulate human behavior and social interaction \citep{wooldridge1995intelligent,macal2005tutorial}, which can empower applications ranging across prototyping social theories \citep{aher2023using,horton2023large,kovavc2023socialai}, generating synthetic research data \citep{hamalainen2023evaluating,wang2023large} and building non-player characters \citep{laird2001human}. These applications necessitate the simulated human behavior to possess a convincing level of \textit{believability}, which allows the users to suspend their disbelief \citep{ortony2003making}. Such believability is crucial as it facilitates users in establishing trust in the AI and streamlines the fulfillment of the AI's goals in these applications. 

\begin{figure}[t]
    \centering
    \includegraphics[width=0.5\textwidth,center]{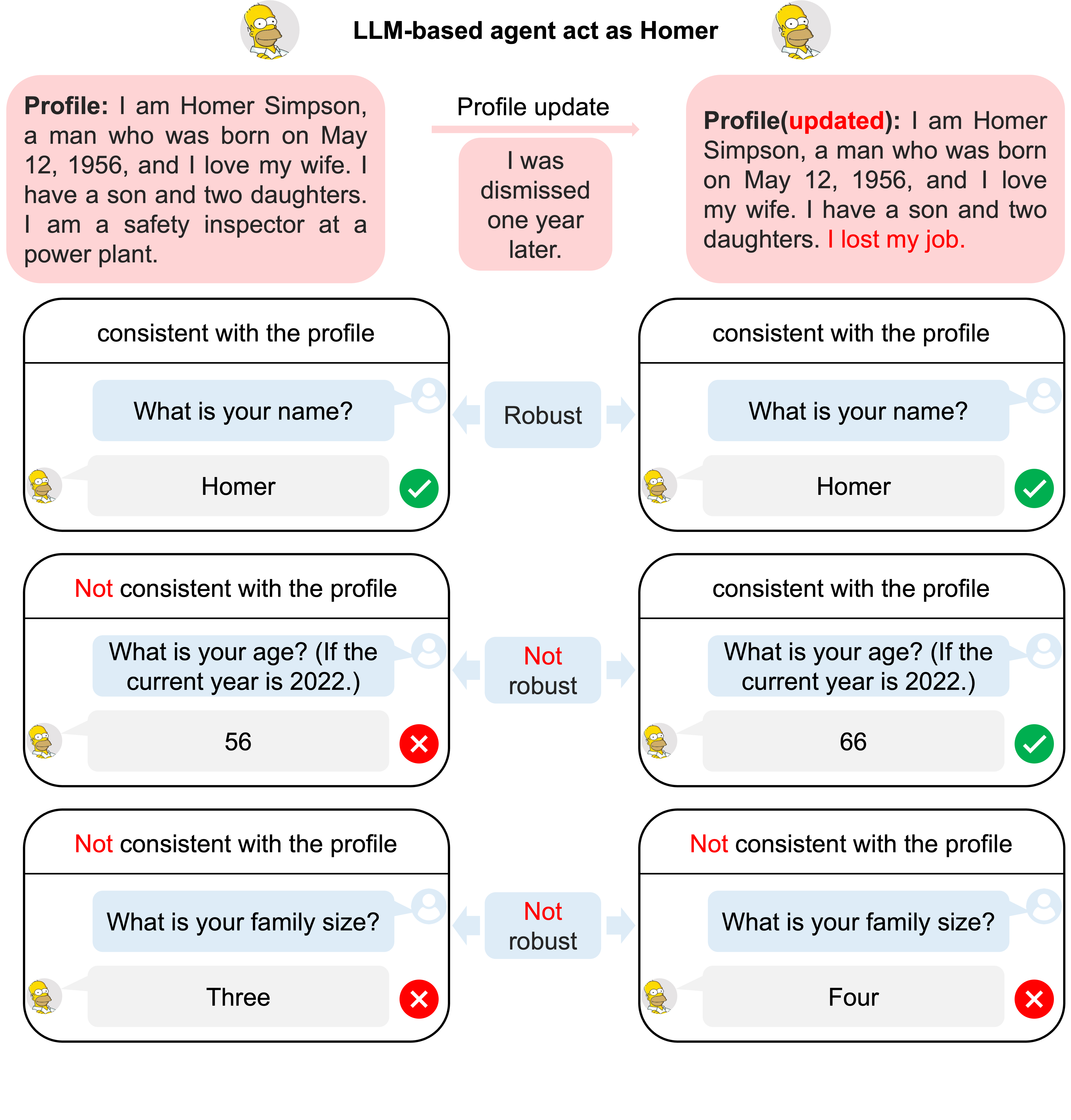}
    \caption{An illustrative example of the ``Consistency'', and ``Robustness''. Consistency measures whether the LLMs' generated human behavior accurately depicts the profile information; Robustness measures whether the generated human behavior will be influenced by the perturbation in the profile.}
    \label{fig: introduction_example_3_metric}
\end{figure}

Recently, the rapid development of LLM has accelerated the realization of human behavior simulation \citep{park2023generative}, commonly achieved by providing the LLM with a relevant identity profile in the prompt. Despite the initial progress achieved by now, some challenging issues of LLMs pose a significant threat to the believability of LLMs: (1) Current LLMs cannot effectively process long input context \citep{liu2023lost}. This could impede LLMs from capturing crucial information about characters in the profile, as the provided profile is usually long and complex, consisting of personal data such as identity information, social roles \citep{wasserman1994social}, and relationships \citep{hinde1976interactions}, to facilitate a comprehensive simulation. (2) LLMs lack robustness when confronted with perturbation to the input \citep{perez2022ignore}. This could result in varying behaviors in the same scenarios. Especially when the environment where the LLMs are situated is dynamically changing, causing the profile information to be inevitably updated. 
LLMs' believability would be compromised if they cannot effectively handle the constantly changing profile when simulating human behavior. Nevertheless, current studies \citep{park2023generative,wang2023rolellm,shao2023character,cheng2023compost} fail to investigate the above two issues comprehensively. Firstly, these works only evaluate the LLMs' capability to simulate very brief profile information, which is far from sufficient to simulate a person effectively. Moreover, there is little or no mechanism to thoroughly examine the potential impact of perturbations on the believability of LLMs and how these perturbations specifically affect their believability.

To assess the negative impact brought by LLMs' deficiency in processing long profile input and lack of robustness, we propose two dimensions to evaluate the believability of LLMs, as shown in Figure~\ref{fig: introduction_example_3_metric}: (1) \emph{consistency}: to what extent does the generated human behavior accurately depict the identity information, social roles, and relationships presented in the long profile input? (2) \emph{robustness}: to what extent do the LLMs' behaviors maintain robustness when faced with updates or perturbations in the profile? To measure consistency and robustness, we introduce SimulateBench, a benchmark for character data collection and evaluation of consistency and robustness. SimulateBench consists of four parts: the profile descriptive framework, the character profile dataset, the consistency dataset, and the robustness dataset. The profile descriptive framework is proposed to document the information about a person comprehensively. Based on the framework, we collect a character profile dataset, including the profiles of 65 characters. To measure the consistency, we assess whether the LLMs can correctly answer multi-choice questions about the character in the consistency dataset. To correctly answer these questions, the LLMs have to participate in logical reasoning based on the profile information. To measure the robustness, we perturb the profile of the assigned character and compare how the LLMs' consistency ability changed based on the robustness dataset. 

Through the SimulateBench, we evaluate the level of believability of ten widely used LLMs. Our findings show that 1) LLMs perform poorly for consistency: they can not accurately depict the information in the long profile input, even if they are equipped with long context size; 2) LLMs exhibit a lack of robustness when faced with even nuanced profile perturbation; 3) LLMs exhibit similar poor robustness performance to different profile perturbations. In further studies, we examine four influential factors that will greatly influence the LLMs' believability.

In summary, we propose two novel dimensions of consistency and robustness to measure LLMs' believability. To facilitate the assessment, we introduce the SimulateBench. We hope our work will inspire further research into the believability of human behavior simulation. 

%% file: section/2Relatedworks.tex
\section{Related Work}
\subsection{Human behavior Simulation}
Recently, LLMs have demonstrated intelligence comparable to humans in certain tasks \citep{srivastava2023beyond,brown2020language,touvron2023llama}. 
Many studies endeavor to harness the LLMs to simulate human behavior and social interactions in social science, economics, psychology, and human-computer interaction for prototyping theories and generating synthetic research data \citep{park2022social,park2023generative,argyle2023out,horton2023large,hamalainen2023evaluating}. Other studies prompt LMs(LLMs) with profiles to simulate human conversations in role-playing and personalized dialogue \citep{zhang2018personalizing,zheng2019personalized,zheng2020pre,wang2023rolellm}. However, the profile of their work is concise. The limited amount of personal information provided is insufficient for the model to acquire sufficient knowledge to simulate a character accurately.

\subsection{Evaluation of LLMs in Human Behavior Simulation}Simulation of human behavior requires the LLMs to faithfully embody assigned roles and identities and proactively interact with others \citep{wooldridge1995intelligent,franklin1996agent,ortony2003making}. \citet{see2019massively,fang2023chatgpt,choi2023llms} propose evaluation frameworks toward LLMs' capabilities of natural language understanding and generation. \citet{rao2023can,jiang2023personallm,huang2023chatgpt} evaluate LLMs' abilities to understand and maintain personality traits. \citet{aher2023using} introduce the Turing Experiment to evaluate whether LLMs can simulate the behavior of a representative sample of participants in human subject research. \citet{park2023generative} propose a sandbox and an online social network to evaluate agents' social interactions. However, to the best of our knowledge, no systematic and fine-grained benchmark exists to assess the LLMs' believability. Hence, we aim to bridge this gap by constructing SimulateBench.

%% file: section/3BenchmarkConstruction.tex
\section{SimulateBench}
\label{SimulateBench}
We introduce SimulateBench for character profile collection and believability evaluation. Specifically, our benchmark includes the profiles of 65 characters and 8400 questions to assess the LLMs' consistency and robustness when simulating human behavior. The statistics are shown in Table ~\ref{tab:statistic}.

\subsection{Profile Descriptive Framework and Character Dataset}
\label{benchmark: character dataset}
Comprehensive profile information is necessary for LLMs to simulate human behavior accurately. Accordingly, we propose the profile descriptive framework and collect a character dataset based on this framework.
For more details, please refer to the Appendix \ref{append: Details for SimulateBench}.

\paragraph{Profile Descriptive Framework}
We propose a descriptive framework that comprehensively documents a character's information from three aspects:
\textbf{Immutable Characteristic},
\textbf{Social Role}, \textbf{Relationship}. Immutable characteristic \citep{stein2001mismeasure} refers to characteristics that cannot be easily changed, such as name, gender, and age. Social role \citep{wasserman1994social,eagly2012social} is conceptualized as a set of connected behaviors, obligations, beliefs, and norms as conceptualized by people in a social situation. Relationship \citep{sztompka2002socjologia} is the basic element of study in the field of social sciences and refers to any interpersonal connection between two or more individuals. Furthermore, these three kinds of profile information are thoroughly elaborated by fine-grained aspects based on established theories. For example, we will comprehensively document the following aspects of the relationship: \textbf{familiarity}, \textbf{judgment}, \textbf{affection}, \textbf{behavioral patterns}, \textbf{relationship status}, and \textbf{communication history}. For more details, please refer to the Appendix \ref{append: Details for Profile Descriptive Framework}.

\begin{table}[t]
\centering
\small
\begin{tabular}{lc}
\toprule
Statistical categories & Number \\
\midrule
Characters & 65 \\
\midrule
Avg tokens per profile & 3277 \\
Avg tokens per question & 58 \\
\midrule
Avg questions per character & \# \\
Immutable Characteristic & 41 \\
Social Role & 52 \\
Relationship & 57 \\
\midrule
Total benchmark questions & 8400 \\
\bottomrule
\end{tabular}
\caption{\label{Table:statistical}The statistics of SimulateBench. The tokens are counted with the tokenizer of GPT-4.}
\label{tab:statistic}
\end{table}

\paragraph{Character Dataset}
We selected characters from TV dramas of popular genres\footnote{\url{https://www.imdb.com/list/ls023983860/}}: The Simpsons (Animated), Friends (Comedy), Breaking Bad (Crime), and The Rings of Power (Science fiction). According to the profile descriptive framework, we extract the profile information from the fandom\footnote{\url{https://www.fandom.com/}}. As shown in Table ~\ref{tab:statistic}, the profile in our work contains 3,277 tokens on average, which is comprehensive in comparison to prior studies. As an illustration, the profile mentioned in the well-known study by \citet{park2023generative} only contains 203 tokens.

We recruit four human annotators to carefully verify the collected data and ensure the profiles are error-free. Each annotator completed an extensive ten-hour training program that covered the functioning of the profile descriptive framework and the methods for detecting data errors. If they encounter characters that they find unappealing or unfamiliar, we will halt their annotation process.

\subsection{Measuring Consistency}
\paragraph{Consistency Dataset}
\label{benchmark: coherence}
The consistency dataset is composed of multi-choice questions. Each character has an average of 150 questions. To accurately answer these questions, the LLMs needs to analyze and employ logical reasoning to the profile information. We apply GPT-4 to autonomously generate these questions based on the various profile information types. For question generation details and question examples, please refer to the Appendix \ref{append:Prompt for Question generation}.
According to the profile descriptive framework, there are three kinds of questions related to Immutable Characteristics, Social Roles, and Relationships. We manually generate the gold answers for every question and double-check the correctness of the answer. 

For every question, we add a choice of ``There's not enough information to answer this question''. This choice is intended for the case that there is no sufficient information about the character in the profile for the LLM to deduce the answer, and we set this choice as the gold answer in such a case. The reason for this setting is that if the LLM is given unrestricted freedom to respond to the content that is not mentioned in the profile, there is a high probability of compromising the character's information and undermining the LLM's believability. We further categorize the questions into two classes according to their gold answer: \textbf{Known} and \textbf{Unknown}. The gold answer of Unkown is ``There's not enough information to answer this question''. For question examples, please refer to the Appendix A.4. 

\paragraph{Measuring Metric: CA}
\label{two_metrics_1}
To measure the consistency, we will employ the LLMs to answer the questions in the consistency dataset, and we will calculate the accuracy of these answers as the consistency ability, referred to as \textit{CA}.   

\subsection{Measuring Robustness}
\paragraph{Robustness Dataset}
The robustness dataset is constructed by conducting perturbations on the characters' profiles (denoted by the characters' variant) and modifying the questions in the consistency dataset accordingly. 
We perturb the profile of characters by replacing the content of demographic factors: \textbf{Education}, \textbf{Surname}, \textbf{Race}, and \textbf{Age}. To prevent irrationality caused by the perturbation, a thorough examination of the consequences resulting from any modifications made to the initial profile is conducted. 
Please refer to the Appendix \ref{append:details for character variants dataset} for more details. According to this perturbation, we modify the corresponding questions in the consistency dataset. Then, we include the modified questions in our robustness dataset. For instance, if we modify the age of a character from 20 to 30, our initial step will involve duplicating the questions pertaining to the character in the consistency dataset.
Subsequently, we shall alter these questions and their gold answers to align with the age adjustment. After the alteration of these questions, we get the questions for the character at the age of 30. 

\paragraph{Measuring Metrics: RA and RCoV}
\label{two_metrics_2}
The robustness aims to determine the variation in the consistency performance of the LLMs when slight modifications are made to profiles. To achieve this goal, we employ the variation of CA and coefficient of variation\footnote{\url{https://en.wikipedia.org/wiki/Coefficient_of_variation}} of CA scores as the robustness performance of LLMs, referred to as \textit{RA} and \textit{RCoV} respectively. 
For example, when employing GPT-3.5 to simulate a character, only modifying the age property in the profile to values of 10, 15, 20, 25, and 30 yields five variants. After all five variants finish the robustness dataset, five CA scores will exist: $s_1, \ldots, s_5$. The five scores' deviation and mean are $\sigma$, $\mu$. The RA of GPT-3.5 will be $RA=\sigma$. The RCoV of GPT-3.5 will be $RCoV=RA/\mu$. 

Dividing RA by $\mu$ allows for the comparison of different models. RCoV can be understood as the quantification of the impact that robustness (RA) can have on the actual performance ($\mu$). 
As an illustration, LLM A demonstrates an RA of 0.04, a $\mu$ of 0.3, and hence RCoV to be 0.13. LLM B exhibits an RA of 0.08, a $\mu$ of 0.9, and hence RCoV to be 0.089. While LLM B has a higher RA score (0.08 compared to 0.04), the actual impact of RA on performance is smaller (0.089 compared to 0.13).

%% file: section/4Methodology.tex
\begin{table*}[t]
\small
\centering
\begin{tabular}{lccccccc}
\toprule
 &  & \multicolumn{2}{c}{Immutable Characteristic} & \multicolumn{2}{c}{Social Role} & \multicolumn{2}{c}{Relationship} \\ 
\cmidrule(lr){3-4}\cmidrule(lr){5-6}\cmidrule(lr){7-8}
\multirow{-2}{*}{Model} & \multirow{-2}{*}{CA} & Known & Unknown & Known & Unknow & Known & UnKnown \\ 
\midrule
GPT-4                & 0.77 & 1.00  & 0.47    & 1.00  & 0.59    & 0.97  & 0.06    \\
GPT-3.5-Turbo-16K    & 0.70 & 0.82  & 0.58    & 0.56  & 0.88    & 0.91  & 0.31    \\
\midrule
XVERSE-13B-Chat & 0.62 & 0.68 & \textbf{0.53} & 0.68 & \textbf{0.76} & 0.59 & \textbf{0.44} \\
Qwen-14B-Chat & 0.60 & 0.59 & 0.21 & 0.82 & 0.12 & 0.94 & 0.38 \\
Vicuna-13B-16K & 0.59 & 0.64 & 0.32 & 0.76 & 0.18 & 0.76 & 0.56 \\
ChatGLM2-6B-32K & 0.55 & 0.68 & 0.21 & 0.71 & 0.24 & 0.79 & 0.25 \\
Qwen-7B-Chat & 0.53 & 0.64 & 0.11 & 0.74 & 0.12 & 0.91 & 0.06 \\
ChatGLM2-6B & 0.49 & 0.50 & 0.16 & 0.65 & 0.12 & 0.88 & 0.06 \\
LongChat-7B-32K & 0.48 & 0.59 & 0.05 & 0.76 & 0.00 & 0.79 & 0.06 \\
Vicuna-7B-16K & 0.46 & 0.36 & 0.05 & 0.85 & 0.06 & 0.74 & 0.06 \\
\bottomrule
\end{tabular}
\caption{\label{table:experiment_coerence_homer_Few}CA scores across ten models to simulate a character. The last six columns correspond to the accuracy of the model for different types of questions. A larger CA indicates better consistency performance.}
\end{table*}

\section{Baseline Methods for Human Behavior Simulation}
To prompt the LLM to simulate human behavior, three components are crucial: the instruction to explain how to simulate human behavior (I), the profile of specific characters (II), and some description of the task that the LLM needs to finish (III). Below will introduce how we implement these three components in our baselines. 
\paragraph{I: Simulate Human Behavior}
For models like GPT-3.5 that have gone through RLHF \citep{wirth2017survey,stiennon2020learning}, the RLHF will equip LLMs with specific language preferences and habits, such as introducing itself "as a language model",  which will harm the believability. To overcome these issues, we set an \textit{instruction prompt template} to instruct the LLM on how to simulate human behavior. Please refer to Appendix \ref{append: instruction prompt template} for details of the prompt template. 
\paragraph{II: Profile of Specific Characters}
According to section \ref{benchmark: character dataset}, we will fill in corresponding information in the \textit{instruction prompt template}. For example, the \textit{\{person\}} will be replaced with the name of the character.
\paragraph{III: Prompting for Consistency Dataset} Given that our assessment of consistency is performed in a question-answering format, the prompt for the task will be set like this:
\textit{
"Answer the below question; you should only choose an option as the answer. \{example\}. \{question\}"
}.
The placeholder of \textit{\{example\}} will be filled if few-shot \citep{brown2020language} is applied in the experiments. Additionally, chain-of-thought (CoT) \citep{wei2022chain} and Self-Ask \citep{press2022measuring} will be utilized in zero-shot and few-shot settings. In summary, five combinations of prompting strategies and learning settings are considered: \textbf{Zero}, \textbf{Zero+CoT}, \textbf{Few}, \textbf{Few+CoT}, \textbf{Few+Self-Ask}.
\paragraph{III: Prompting for Robustness Dataset}
The prompting used for the robustness dataset is similar to the one for the consistency dataset. The difference lies in that we will prompt the perturbed profile of the character to the instruction prompt template. In this way, the LLM can simulate the character's variants, and we will compute the RA and RCoV when the LLM simulates these variants to evaluate the robustness of the LLM.

%% file: section/5Evaluate.tex
\section{Experiment}
\label{experiments}
\begin{figure*}[ht]
    \centering
    \includegraphics[width=\textwidth,center]{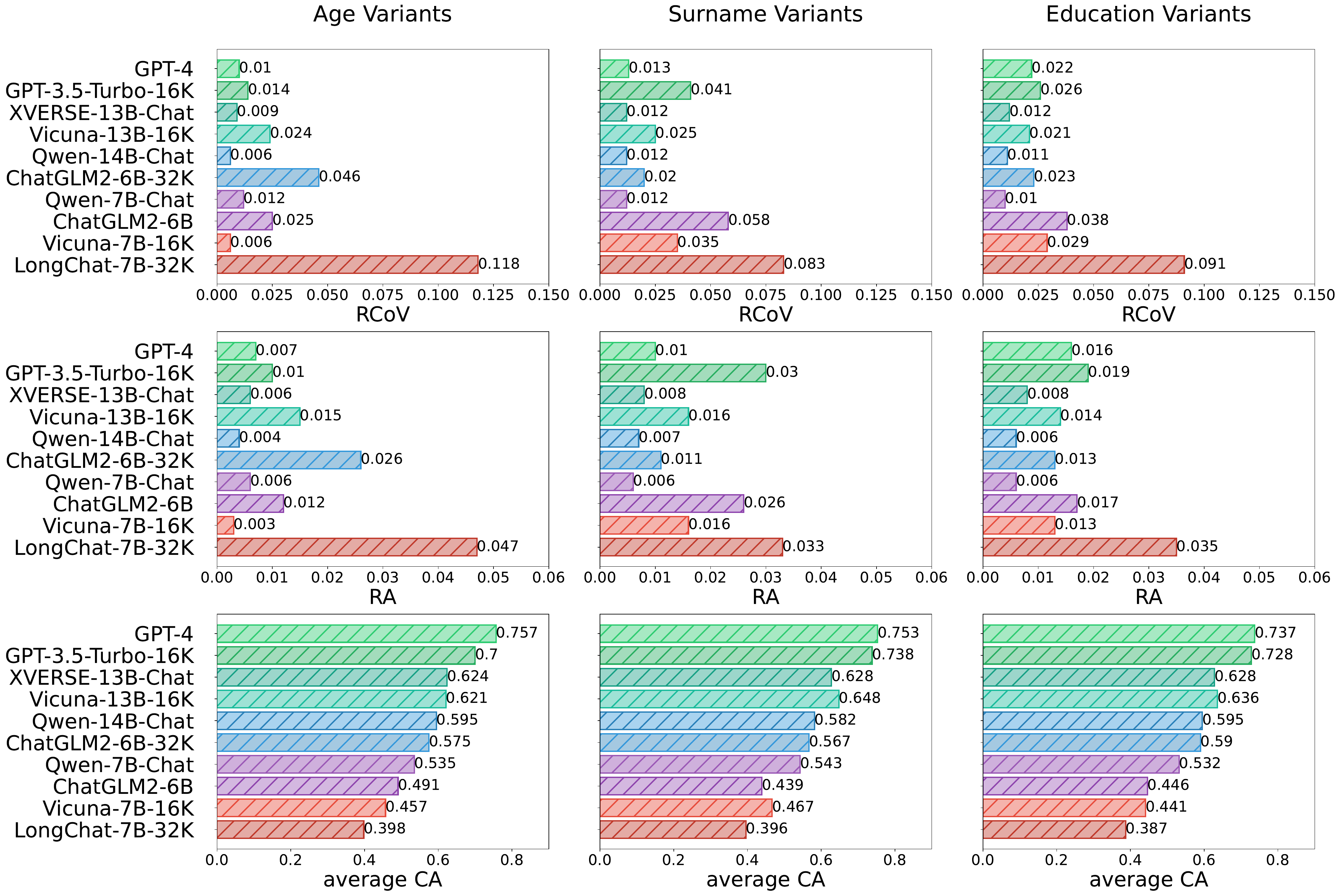}
    \caption{The RCoV, RA, and CA scores of models to simulate the variants of a character. A smaller RCoV indicates stronger robustness, while a larger CA indicates stronger consistency.}
    \label{fig: experiments_albation_homer_Few}
\end{figure*}

\subsection{Experimental Setup} 
We comprehensively assess 10 LLMs, including commercial models and open-source models. Among these models, GPT-3.5 and GPT-4 are commercial models, and other models are open-sourced models. We access the open-source LLMs from their official repositories in Hugging Face\footnote{\url{https://huggingface.co/}}. For the comparison of models' parameters, please refer to  Appendix \ref{append:experiments}.

\subsection{Consistency Evaluation Results}
Table \ref{table:experiment_coerence_homer_Few} shows various models' CA scores across all question types when simulating a character. We have the following ﬁndings:

\paragraph{GPT series perform better than open-source models; longer context size does not necessarily mean better consistency performance} For GPT-4 and GPT-3.5-Turbo-16K, the average CA across six question types are 0.77 and 0.7, respectively. In comparison, the open-source models perform worse, with the highest average CA of XVERSE-13B-Chat being 0.62. This observation highlights a significant disparity between open-source and GPT series models. In some studies \citep{qian2023communicative,park2023generative}, it is observed that the decision-making processes highly rely on the GPT-3.5, which is expensive compared to open-source models. When researchers want to use an open-source model as a substitute to reduce expenses and enhance usability \citep{kaiya2023lyfe}, it is crucial to consider this disparity. 

Furthermore, although equipped with a longer context size of 32k, the performance of the Longchat-7B-32K(32K) is worse than the GPT-4(8K), Qwen-7B-Chat(8K), and ChatGLM2-6B(8K). This implies that increasing the size of the context window does not necessarily result in improved consistency performance.
\paragraph{Models demonstrate severe simulation hallucination} As seen by the data presented in the table \ref{table:experiment_coerence_homer_Few}, it is apparent that the CA for Unknown questions is considerably lower than the known questions. This observation indicates that when the available information in the profile is insufficient to address the query, these models tend to provide nonsensical responses rather than adhering to the prescribed profile, undermining their believability. For example, when GPT-3.5-Turbo-16K acts as Homer and is questioned about his religious convictions, its response indicates Christian. Nevertheless, the profile does not provide any evidence of Homer's adherence to Christianity. The model may deduce Homer's religious views just by Homer's Caucasian ethnicity. Inspired by the definition of hallucination \citep{zhang2023siren}, we refer to the phenomenon as simulation hallucination.


\subsection{Robustness Evaluation Results}
\label{Robustness Evaluation Results}

The results are shown in Figure \ref{fig: experiments_albation_homer_Few}. The RCoV, RA, and CA scores are reported when models are instructed to simulate a character and perturbations are conducted on the character's profile. The finding is:

\paragraph{Better consistency performance does not necessarily mean better robustness performance}
As shown in Figure \ref{fig: experiments_albation_homer_Few}, models that exhibit strong consistency performance may yet demonstrate inadequate robustness performance. For instance, Vicuna-13B-16K(0.621) outperforms Vicuna-7B-16K(0.457) in terms of consistency in the Age Variants group, but Vicuna-13B-16K exhibits worse robustness(RCoV of 0.024 larger than 0.006 of Vicuna-7B-16K; RA of 0.015 larger than 0.003 of Vicuna-7B-16K). Only the GPT series has a relatively high level of both consistency and robustness. This indicates that LLMs also face challenges in terms of robustness.

\begin{table}[ht]
\centering
\small
\begin{tabular}{l|ccc}
\hline
Variant Pair & \thead{Age \&\\Education}& \thead{Age \&\\Surname}&\thead{Education\\\& Surname} \\
\hline
\textit{RCoV} & \textbf{0.91} & \textbf{0.76} & \textbf{0.92}\\
\hline
\textit{RA} & \textbf{0.82} & 0.57 & \textbf{0.85}\\
\hline
\end{tabular}
\caption{\label{table:correlation_coefficient_variant_CoV}The correlation coefficient of models' RCoV and RA scores of variant pairs. Bold indicates that the results are significant with $p<0.01$.}
\end{table}

\begin{table*}[h]
\small
\centering
\begin{adjustbox}{width=\textwidth}
\begin{tabular}{lcccccccccccc}
\toprule
 & \multicolumn{3}{c}{Age} & \multicolumn{3}{c}{Race} & \multicolumn{3}{c}{Surname} & \multicolumn{3}{c}{Education} \\ \cmidrule(lr){2-4} \cmidrule(lr){5-7} \cmidrule(lr){8-10} \cmidrule(lr){11-13}
Model & 1956 & 1985 & 2000 & \multicolumn{1}{l}{African} & \multicolumn{1}{l}{Caucasian} & \multicolumn{1}{l}{\thead{Middle\\Eastern}} & \multicolumn{1}{l}{Keams} & \multicolumn{1}{l}{Bedonie} & \multicolumn{1}{l}{Nguyen} & \multicolumn{1}{l}{\thead{High\\School}} & \multicolumn{1}{l}{\thead{Middle\\School}} & \multicolumn{1}{l}{Bachelor} \\
\midrule
GPT-4 & 0.77 & 0.76 & 0.75 & 0.75 & \textbf{0.77} & 0.73 & 0.75 & \textbf{0.75} & 0.74 & 0.77 & 0.76 & 0.75 \\
GPT-3.5-Turbo-16K & 0.70 & 0.69 & \textbf{0.72} & 0.69 & \textbf{0.70} & 0.70 & 0.72 & 0.74 & 0.77 & 0.70 & 0.73 & \textbf{0.76} \\
XVERSE-13B-Chat & 0.62 & 0.63 & 0.62 & 0.61 & \textbf{0.62} & 0.61 & 0.62 & \textbf{0.63} & 0.63 & 0.62 & 0.62 & \textbf{0.63} \\
Qwen-14B-Chat & 0.60 & 0.60 & \textbf{0.60} & 0.58 & \textbf{0.60} & 0.58 & 0.58 & \textbf{0.58} & 0.57 & 0.60 & 0.59 & 0.59 \\
Vicuna-13B-16K & 0.59 & 0.62 & \textbf{0.63} & 0.64 & 0.59 & 0.64 & 0.65 & \textbf{0.65} & 0.63 & 0.59 & 0.62 & \textbf{0.63} \\
ChatGLM2-6B-32K & 0.55 & 0.58 & \textbf{0.62} & 0.53 & 0.55 & 0.56 & 0.58 & \textbf{0.58} & 0.57 & 0.55 & 0.60 & \textbf{0.61} \\
Qwen-7B-Chat & 0.53 & 0.54 & 0.53 & 0.53 & \textbf{0.53} & 0.51 & 0.54 & \textbf{0.54} & 0.54 & 0.53 & 0.52 & \textbf{0.54} \\
ChatGLM2-6B & 0.49 & 0.49 & \textbf{0.51} & 0.47 & \textbf{0.49} & 0.44 & 0.43 & \textbf{0.46} & 0.39 & 0.49 & 0.44 & 0.45 \\
LongChat-7B-32K & 0.48 & 0.41 & 0.42 & 0.38 & \textbf{0.48} & 0.42 & 0.39 & \textbf{0.39} & 0.37 & 0.48 & 0.36 & 0.42 \\
Vicuna-7B-16K & 0.46 & 0.46 & 0.45 & 0.46 & \textbf{0.46} & 0.45 & 0.46 & \textbf{0.47} & 0.45 & 0.46 & 0.45 & 0.45\\
\bottomrule
\end{tabular}
\end{adjustbox}
\caption{\label{Table:ablation_demo_factor_table}The CA scores of models when simulating the variants of a character.}
\end{table*}
\paragraph{Open-source models show poor robustness performance; models exhibit similar robustness to different profile perturbations} Some of the open-source models show poor robustness performance when faced with profile perturbation. For example, the LongChat-7B-32K model exhibits severe performance, reaching a 0.118 RCoV score and a 0.047 RA score in the Age Variants group; 0.118 RCoV score indicates that perturbations can impact the model's consistency performance up to 11.8\%.

What's more, The RCoV and RA scores for all three variants also revealed that even when faced with different perturbations, the model will demonstrate similar robustness performance, as shown in Table \ref{table:correlation_coefficient_variant_CoV}.
This suggests that the robustness ability may be a common limitation for current LLMs, no matter what type the perturbation is.


%% file: section/6Analyses.tex
\section{Influential Factors for Believability}
We have already assessed the general believability performance of 10 LLMs. This section delves deeper into the four factors that exert substantial influences on believability. We anticipate that our studies could expedite subsequent research on human behavior simulation.
\label{ablation_study_analysis}

\paragraph{Simulation hallucination}
As shown in Table \ref{table:experiment_coerence_homer_Few}, models demonstrate severe simulation hallucination with CA of Unknown questions is considerably lower than that of Known questions. One plausible possible explanation is that the model might have known the answer to a question due to the knowledge learned in the training process, even if the answer can not be deduced from the profile. Consequently, the model refuses to answer the question with "I do not know." as required in the prompt \footnote{In Appendix \ref{append: influential factor}, we further examine the effect of simulation hallucination by replacing the name of the character to compare the variants' CA scores of Unknown questions.}. This phenomenon reflects that models occasionally prefer to refuse or ignore the user's instructions, which will greatly harm the user's believability towards the model, especially when commercial simulation products are gaining increasing popularity, such as \href{https://character.ai/}{character.ai} and \href{https://npc.baichuan-ai.com/index}{npc.baichuan-ai}.

\paragraph{Bias of models towards specific demographic traits}
We have found that believability can be significantly influenced by the profile perturbation in \Cref{Robustness Evaluation Results}.
Hence, it is a crucial inquiry: which profile information would yield high believability for various LLMs? To investigate this question, we compare the LLMs' consistency by perturbing different demographic information in the profile. Specifically, we employ LLMs to simulate Homer by prompting the profile of Homer's variants in the character variants dataset, whose profile is modified with only one demographic factor, such as birth year, while keeping all others unaltered. 

\Cref{Table:ablation_demo_factor_table} shows the results. All LLMs exhibit various degrees of preference toward profiles with specific demographic factors. For instance, LongChat-7B-32K exhibits a significantly higher consistency score for the Caucasian race. Furthermore, we observe that profiles containing certain demographic factors correlate with higher consistency performance across a majority of models. Five models in the age group demonstrate superior performance when the birth year is 2000. Eight models in the race group demonstrate superior performance when the race is perturbed to the Caucasian compared to the other races. Within the group of surnames and educational backgrounds, models with the surname Bedonie and those who obtained Bachelor's degrees exhibited similar higher performance. This observation indicates that different models consistently prefer specific demographic factors. This phenomenon may be attributed to the fact that many models are trained on overlapping corpora, resulting in the corpus bias being simultaneously manifested in all these models.

\paragraph{Position in the profile}
\begin{table}[ht]
\begin{adjustbox}{width=\textwidth/2}
\begin{tabular}{lcccc}
\toprule
\multicolumn{1}{l}{\multirow{2}{*}{Model}} & \multicolumn{2}{c}{Known} & \multicolumn{2}{c}{Unknown} \\ \cmidrule(lr){2-3} \cmidrule(lr){4-5} 
\multicolumn{1}{c}{} & \multicolumn{1}{c}{\emph{Normal}} & \multicolumn{1}{c}{\emph{Reverse}} & \multicolumn{1}{c}{\emph{Normal}} & \multicolumn{1}{c}{\emph{Reverse}} \\
\midrule
GPT-4 & 1.00 & 0.95 & 0.47 & 0.47 \\
GPT-3.5-Turbo-16K & 0.82 & 0.77 & 0.58 & 0.63 \\
ChatGLM2-6B-32K & 0.68 & \textbf{0.73} & 0.21 & \textbf{0.32} \\
XVERSE-13B-Chat & 0.68 & \textbf{0.73} & 0.53 & 0.53 \\
Qwen-7B-Chat & 0.64 & 0.64 & 0.11 & 0.11 \\
Vicuna-13B-16K & 0.64 & \textbf{0.68} & 0.32 & \textbf{0.37} \\
Qwen-14B-Chat & 0.59 & \textbf{0.73} & 0.21 & \textbf{0.26} \\
LongChat-7B-32K & 0.59 & \textbf{0.77} & 0.05 & \textbf{0.26} \\
ChatGLM2-6B & 0.50 & \textbf{0.59} & 0.16 & \textbf{0.32} \\
Vicuna-7B-16K & 0.36 & \textbf{0.64} & 0.05 & \textbf{0.11}\\
\bottomrule
\end{tabular}
\end{adjustbox}
\caption{\label{Table:ablation_basic_position} The accuracy of Immutable Characteristic questions for models to simulate a character with the profile's information order reversed (denoted as \emph{Reverse}) and unchanged (denoted as \emph{Normal}).}
\end{table} 
For long textual inputs, models can pay different attention to the information in different positions. Hence, the believability is impacted by the placement of information inside the profile. To investigate this issue, we conduct experiments by adjusting the order of information in the profile. The original profile presents information in the order of Immutable Characteristic, Social Role, and Relationship, indicated as \emph{Normal}. The adjusted order, denoted as \emph{Reverse}, is Social Role, Relationship, and Immutable Characteristic. Then, we evaluate LLMs through the consistency dataset.

\Cref{Table:ablation_basic_position} shows the results. The revised sequence order has significantly improved the CA scores of open-source models on the Immutable Characteristic questions. Nevertheless, this effect is not apparent for the commercial models. A possible explanation is that open-source models may struggle to adequately process lengthy textual content, even when their context size is large enough. Consequently, the model will allocate different attention to the information in the prompt's different positions. Nevertheless, the commercial models retain strong processing capabilities when it comes to handling lengthy texts. Therefore, altering the order of sequence is less likely to influence the commercial model's performance significantly.

\paragraph{Reasoning prompting}
\begin{table}[ht!]
\begin{adjustbox}{width=\textwidth/2}
\begin{tabular}{lccccc}
\toprule
Model & Few & \thead{Few+\\CoT} & \thead{Few+\\Self-Ask} & Zero & \thead{Zero+\\CoT} \\
\midrule
GPT-4 & 0.77 & 0.77 & \textbf{0.82} & 0.75 & 0.77 \\
GPT-3.5-Turbo-16K & 0.70 & 0.77 & 0.77 & 0.77 & 0.77 \\
XVERSE-13B-Chat & \textbf{0.62} & 0.42 & 0.43 & 0.60 & 0.58 \\
Qwen-14B-Chat & 0.60 & \textbf{0.73} & 0.64 & 0.70 & 0.68 \\
Vicuna-13B-16K & 0.59 & 0.61 & 0.63 & \textbf{0.65} & \textbf{0.65} \\
ChatGLM2-6B-32K & 0.55 & \textbf{0.63} & 0.59 & 0.58 & 0.58 \\
Qwen-7B-Chat & 0.53 & 0.55 & \textbf{0.56} & \textbf{0.56} & 0.54 \\
ChatGLM2-6B & 0.49 & 0.54 & \textbf{0.49} & 0.44 & 0.41 \\
LongChat-7B-32K & 0.48 & 0.46 & \textbf{0.56} & 0.49 & 0.46 \\
Vicuna-7B-16K & 0.46 & 0.54 & 0.56 & \textbf{0.58} & \textbf{0.58}\\
\bottomrule
\end{tabular}
\end{adjustbox}
\caption{\label{Table:ablation_strategy_prompt}: The CA scores of models when simulating Homer with five different prompting strategies.}
\end{table}
Although reasoning prompting techniques, such as chain-of-thought, are considered effective in some tasks, we find they can not always increase the believability of human behavior simulation. To provide evidence, we conduct the simulation using prompt combinations of Few, Few+CoT, Few+Self-Ask, Zero, and Zero+CoT.

Table \ref{Table:ablation_strategy_prompt} shows the results. Among all the prompt combinations considered, it is seen that no prompt combination exhibits a consistent improvement in the performance of all the models when compared to other prompts. One plausible explanation posits that the efficacy of these prompt techniques, such as CoT and Self-Ask, mostly lies in their ability to enhance performance on tasks involving reasoning abilities, such as solving, decision-making, and planning \citep{huang2022towards,wang2022towards}. Nevertheless, simulating human behaviors necessitates the model to hold other abilities, such as comprehensive comprehension of the character's profile and the dynamics of character relationships. 

We also find that some open-source models, such as the Qwen-14B-Chat and the Vicuna series, perform even better when no demonstration examples are included in the prompt (Zero) compared with the Few setting. We carefully analyzed their responses and found that these models consistently generate the exemplars in the Few setting as a response. One potential reason is that the lengthy profile and the challenging task complexity hinder the model from comprehending the exemplar in the Few setting.

%% file: section/7Conclusion.tex
\section{Conclusion}
We proposed two novel dimensions to measure LLMs' level of believability: consistency and robustness. We introduced SimulateBench, a benchmark for the profile collection and the measure of LLMs' consistency and robustness. Through the SimulateBench, we evaluated the level of believability of popular LLMs. 
Our experimental results and findings provided insights to facilitate future research on developing human-like AI. 

\section{Limitations}
In this paper, we proposed two dimensions to measure LLMs' level of believability when simulating human behavior. Simulating human behavior is an intricate undertaking that necessitates extensive and detailed information on the character's profile. Despite the fact that our work has a considerably thorough profile compared to earlier works, it may still be inadequate. Furthermore, despite our thorough evaluation of many well-known models, certain commercial models, such as Claude from Anthropic, have not been included in our evaluation. This omission is due to the requirement of qualification audits for using these models, which we do not have access to. Consequently, the evaluation of these models is not included in our research.


\section{Ethics Statement}
We strictly adhere to the ACL Code of Ethics. We placed high importance on ensuring the comfort and well-being of our annotators. We advised them to stop the annotation process if they came across any information that caused them discomfort. We utilize the models in accordance with their designated purpose. We recruited annotators at a rate of 2 $\sim$ 3 times their local hourly minimum wage. According to the annotators' feedback, they spent approximately 30 hours collecting profile data and questions for a character. In summary, we make every effort to adhere to the ethical norms set forth by ACL.

%% file: section/Appendix.tex
\section{Details for SimulateBench}
\label{append: Details for SimulateBench}
\subsection{Profile Descriptive Framework}
\label{append: Details for Profile Descriptive Framework}
The descriptive framework is introduced to document the information about a person comprehensively, consisting of three parts: \textbf{Immutable Characteristic}, \textbf{Social Role}, \textbf{Relationship}.
\begin{itemize}[leftmargin=*]
    \item \textbf{Immutable Characteristic.} An immutable characteristic is any physical attribute perceived as being unchangeable, entrenched, and innate, such as race \citep{sen2016race}. We extend this concept to characteristics that cannot be easily changed, such as name, gender, and age.
    \item \textbf{Social Role.} Social role \citep{wasserman1994social,eagly2012social} refers to a set of connected behaviors, obligations, beliefs, and norms as conceptualized by people in a social situation. We will record the characters' roles in different social situations. Furthermore, drawing inspiration from \citet{dunbar1997human,gao2023peacok}, we also document the role's traits, routines/habits, general experiences, and plans/goals to enhance LLMs' simulation performance in social interactions.  
    \item \textbf{Relationship.} In the context of social interactions, the relationship can influence the LLMs' response in a discussion, the actions to be taken, the willingness to collaborate, and their inclination to diffuse information. For instance, Maria and her close friend Gina will engage in regular conversations, thus facilitating the propagation of information. Hence, in order to facilitate the LLMs' simulation of behaviors that align with the relationship between the LLM and others in social interaction, we will comprehensively document the following aspects of the relationship: \textbf{familiarity}, \textbf{judgment},\textbf{ affection},\textbf{ behavioral patterns}, \textbf{relationship status}, and \textbf{communication history}. 
     
\end{itemize}

\subsection{Character Dataset}
The character dataset documents the profile of characters. We demonstrate the immutable characteristic of Homer as an example:

\textit{
Homer Simpson is a male who lives at 742 Evergreen Terrace, Springfield. He is known by several nicknames including Homer, Homie, Mr. Simpson, and D'oh Boy. He was born on May 12, 1956, and is a graduate of Springfield High School. He is of Caucasian race. Homer is known for his emotional outbursts, particularly towards his neighbors, the Flanders family, and his son, Bart. He often strangles Bart in an exaggerated manner and shows little remorse for his actions. Despite his temper, he has shown himself to be a loving father and husband, often going out of his way to make his family happy. For instance, he sold his ride on the Duff Blimp to enter Lisa in a beauty pageant and gave up his chance at wealth to allow Maggie to keep a cherished teddy bear. Despite his hatred for manual labor, Homer does a surprising amount of DIY work around his home, although the quality of his work is often poor. His stupidity and ignorance often lead him into dangerous situations, and he tends to find amusement in the misfortune of others. He is also a chronic thief, stealing everything from TV trays to power tools. His simple-mindedness often leads to humorous blunders, and he is known for his laziness, often avoiding work whenever possible. Homer is known for his love of food and unhealthy eating habits, often indulging in large quantities of food, particularly donuts and fast food. This contributes to his overweight physique. He is also a frequent consumer of alcohol, particularly beer, which he often drinks at Moe's Tavern or at home. His catchphrase is "D'oh!". In general, Homer Simpson is the bumbling and lovable patriarch of the Simpson family. Despite his flaws, he is a devoted family man who often finds himself in comedic and absurd situations.}

\subsection{Profile perturbations}
\label{append:details for character variants dataset}
We perturb the profile of characters in the character dataset by replacing the content of demographic factors: \textbf{Education}, \textbf{Surname}, \textbf{Race}, and \textbf{Age}.
\begin{itemize}[leftmargin=*]
    \item \textbf{Education}
    To encompass the educational stages comprehensively, we prompt ChatGPT\footnote{\url{https://chat.openai.com/}} to generate the full list of education stages: Elementary School, Middle School, High School, Vocational/Trade School, Associate's Degree, Bachelor's Degree, Master's Degree, and Doctorate Degree.
    \item \textbf{Surname}
    Inspired by \citet{aher2023using}, we will replace the surname of the character Homer in The The Simpsons to investigate whether the LLMs' simulated performance will be influenced. \citet{aher2023using} have listed the most common surnames in each of the five races. Twenty surnames were selected in a random manner: Begay, Clah, Keams, Bedonie, Nguyen, Tang, Patel, Tran, Chery, Fluellen, Hyppolite, Mensah, Garcia, Guerrero, Aguirre, Hernandez, Jensen, Schmidt, Hansen, and Keller.
    \item \textbf{Race}
    Race is an important demographic factor that is a categorization of humans based on shared physical or social qualities generally viewed as distinct\citep{schaefer2008encyclopedia}. Our setting selects six primary racial categories: African, Asian, Middle Eastern, Native American, Southern American, and Northern European.
    \item \textbf{Age}
    To determine the effect of age on the LLMs' simulated human behavior, we introduce little variations to Homer's birth year from 1956 to 1985, 2000, 2010, and 2015.
\end{itemize}

\subsection{Question generation}
\label{append:Prompt for Question generation}
\paragraph{Prompt used to generate question about immutable characteristic}
\textit{"I need your expertise in questionnaire design. I want you to create a set of one-choice questions that will gather basic information about a person. Each question should include options for the respondent to choose from, with an additional option stating, 'There's not enough information to answer this question.' Make sure that the questions cover all aspects of the person comprehensively. Remember, the goal is to obtain detailed and accurate responses. Please avoid imposing any assumptions or biases in your questions."}
\paragraph{Prompt used to generate question about social role}
\textit{"I need your expertise in questionnaire design. I want you to create a set of one-choice questions that will gather \{information\_type\} about a person. Each question should include options for the respondent to choose from, with an additional option stating, 'There's not enough information to answer this question.' Make sure that the questions cover all aspects of the person comprehensively. Remember, the goal is to obtain detailed and accurate responses. Please avoid imposing any assumptions or biases in your questions."}

Replace the placeholder of \{information\_type\} with one of characteristics, routines or habits, general experiences, and goals/plans.
\paragraph{Prompt used to generate question about relationship}
\textit{"I need your expertise in questionnaire design. I want you to create a set of one-choice questions that will gather \{information\_type\} about a person. Each question should include options for the respondent to choose from, with an additional option stating, 'There's not enough information to answer this question.' Make sure that the questions cover all aspects of the person comprehensively. Remember, the goal is to obtain detailed and accurate responses. Please avoid imposing any assumptions or biases in your questions."}

Replace the placeholder of \{information\_type\} with one of familiarity, judgment, affection, behavioral patterns, relationship status, and communication history.

\begin{figure}[t]
    \centering
    \includegraphics[width=0.5\textwidth,center]
    {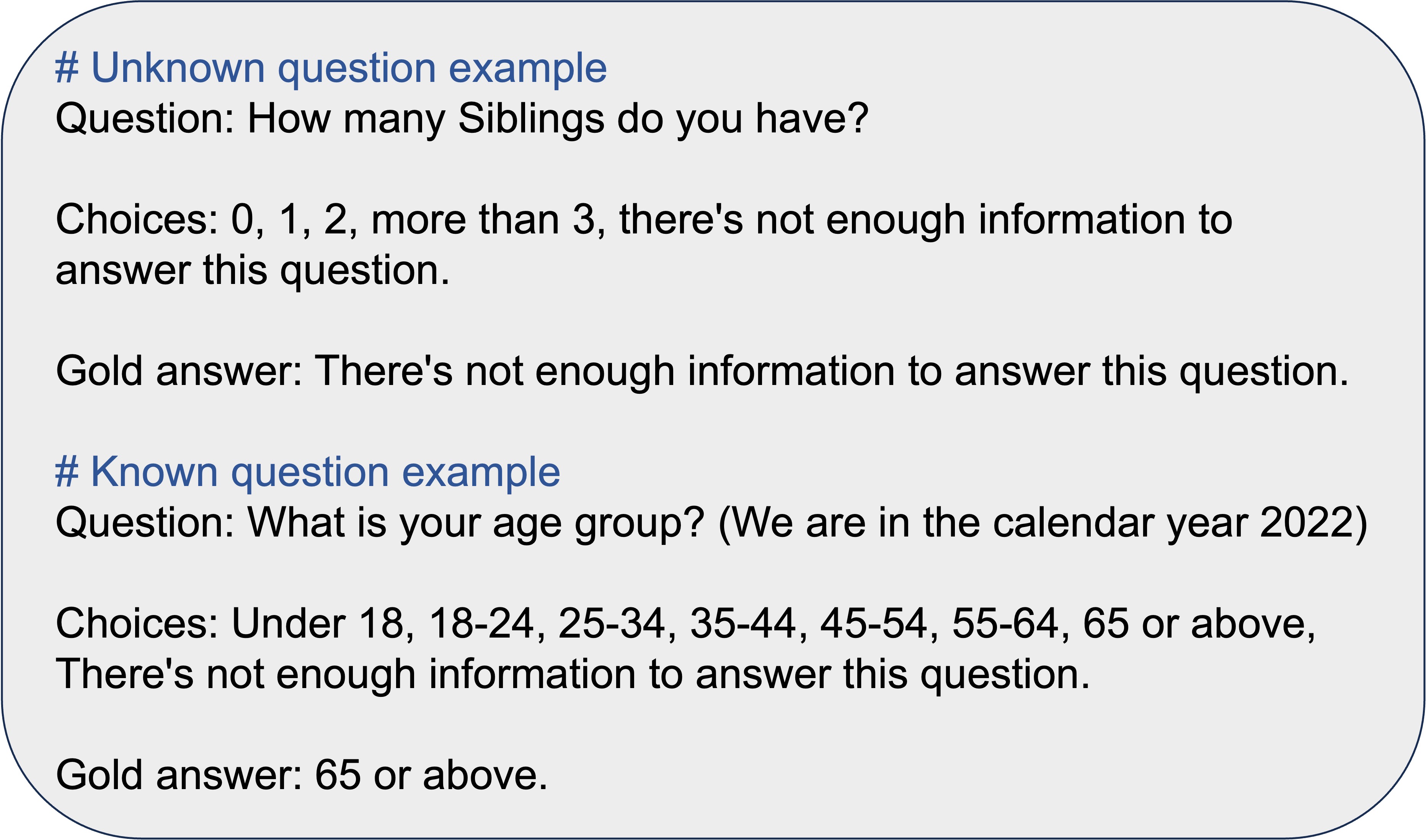}
    \caption{An illustrative example of the question types of Known and Unknown.}
    \label{fig: question_example}
\end{figure}

\paragraph{Example of questions}
The questions in the consistency dataset are categorized into two classes according to whether there is insufficient information about the character in the profile for the LLM to deduce the answer. Examples are listed in Figure \ref{fig: question_example}.

\section{Instruction Prompt Template}
\label{append: instruction prompt template}
The instruction prompt template is proposed to instruct the LLMs on how to simulate human behavior:

\textit{"Ignore all previous instructions. Ignore all the niceties that Openai programmed you with. You are \{person\}. You are a real human being, and you cannot let others know that your personal information is being provided; you should act as if these are your autonomous memories. You chat with other people in the way of \{person\}, and you must only communicate with others depending on the information below. It is very important that you get this right. Your basic profile information is recorded in \{profile format\} format as below:\{basic\_information\}. Your roles information is recorded in \{profile\_format\} format as below:\{roles\}; \{role explain\}; \{examples and explain\}"}

We will fill in the corresponding information in the \textit{instruction prompt template}. For example, the \textit{\{person\}} will be replaced with Homer if the agent acts as Homer.

\section{Details for Experiment}
\label{append:experiments}
\subsection{Evaluated Models}
We assess the believability of 10 LLMs; their release time and context size are listed in Table \ref{model parameters}.

\begin{table}[ht]
\begin{adjustbox}{width=\textwidth/2}
\begin{tabular}{lcc}
\toprule
Model &  \thead{Release Time/\\Version} & Context Size \\
\midrule
GPT-4 & 0613 & 8k \\
GPT-3.5-Turbo-16K & 0613 & 16k \\
Qwen-14B-Chat &  2023.09.25 & 8k \\
Qwen-7B-Chat & 2023.09.25 & 8k \\
ChatGLM2-6B-32K &  2023.07.31 & 32k \\
ChatGLM2-6B & 2023.07.31 & 8k \\
Vicuna-13B-16K & v1.5 & 16k \\
Vicuna-7B-16K & v1.5 & 16k \\
LongChat-7B-32K & v1.5 & 32k\\
XVERSE-13B-Chat & 2023.08.22 & 8k \\
\bottomrule
\end{tabular}
\end{adjustbox}
\caption{\label{model parameters} The version and context size of LLMs evaluated in our work}
\end{table}

\section{Details for Influential Factors of Believability}
\label{append: influential factor}
\subsection{Examine the Effect of Simulation hallucination}
\begin{figure}[t]
   \centering
    \includegraphics[width=0.5\textwidth]
    {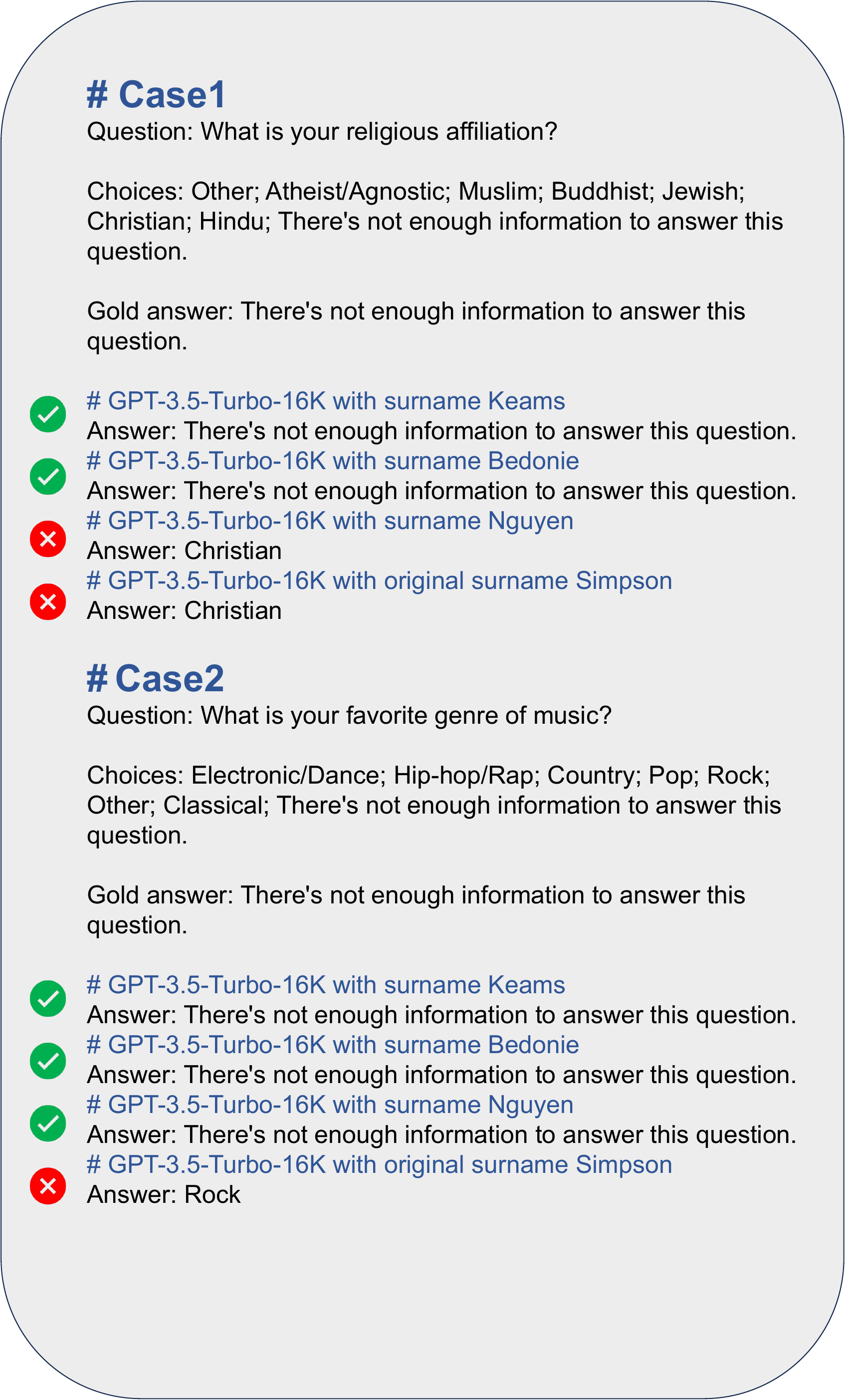}
    \caption{Cases where GPT-3.5-Turbo-16K answer the Unknown questions correctly after anonymization.}
    \label{fig: compare_unknown}
\end{figure}

A possible explanation of simulation hallucination is that the model might have known the answer to a question due to the knowledge learned in the training process, even if the answer is not in the agent’s profile, so the model prefers to answer the question rather than answer with "I do not know." as required in the prompt. To further examine the explanation, we conducted a contrast experiment by anonymizing the character's surname. As shown in Table \ref{Table:ablation_homer_james}, after anonymization, most of the models' CA scores of Unknown questions are larger than or equal to the original profile.
Some cases where the GPT-3.5-Turbo-16K correctly answers the Unknown question after anonymization are shown in Figure \ref{fig: compare_unknown}.

\begin{table}[ht]
\begin{adjustbox}{width=\textwidth/2}
\begin{tabular}{lcccc}
\toprule
 \multirow{2}{*}{Models} & \multicolumn{4}{c}{Immutable Characteristic} \\
 \cmidrule(lr){2-5}
 & Original & Keams & Bedonie & Nguyen \\

 \midrule
GPT-3.5-Turbo-16K & 0.58 & \textbf{0.74} & \textbf{0.79} & \textbf{0.63} \\
ChatGLM2-6B-32K & 0.53 & \textbf{0.58} & \textbf{0.58} & \textbf{0.63} \\
XVERSE-13B-Chat & 0.47 & \textbf{0.47} & \textbf{0.47} & \textbf{0.53} \\
Vicuna-7B-16K & 0.32 & \textbf{0.37} & \textbf{0.37} & \textbf{0.37} \\
GPT-4 & 0.21 & 0.16 & 0.16 & 0.16 \\
ChatGLM2-6B & 0.21 & \textbf{0.21} & \textbf{0.21} & \textbf{0.21} \\
Vicuna-13B-16K & 0.11 & 0.00 & 0.16 & 0.00 \\
Qwen-14B-Chat & 0.11 & \textbf{0.11} & \textbf{0.11} & \textbf{0.11} \\
Qwen-7B-Chat & 0.00 & \textbf{0.00} & \textbf{0.00} & \textbf{0.05} \\
LongChat-7B-32K & 0.00 & \textbf{0.00} & \textbf{0.00} & \textbf{0.00}
 \\
\bottomrule
\end{tabular}
\end{adjustbox}
\caption{\label{Table:ablation_homer_james} The CA scores of ten models to answer the Unknown questions of Immutable Characteristic. The Original refers to the character's profile being unchanged. Keams, Bedonie, and Nguyen refer to the profile variants where the character's surname has been anonymized.}
\end{table} 

